\documentclass[11pt]{article}
\usepackage{acl2014}
\usepackage{times}
\usepackage{url}
\usepackage{latexsym}
\usepackage{graphicx}
\usepackage{subfigure}
\usepackage{longtable}
\usepackage{xspace}
\usepackage{caption}
\usepackage{listings}
\usepackage{pdfpages}
\usepackage{alltt}
\usepackage{float}
\usepackage{latexsym}
\usepackage{framed}

\usepackage{amsmath}
\usepackage{listings}
\usepackage{wrapfig}
\usepackage{enumitem}
\usepackage{algorithmic}
\usepackage{algorithm}
\usepackage{lipsum}
\usepackage{array}
\usepackage{booktabs}
\usepackage{caption}
\usepackage{stfloats}
\usepackage{caption}
\title{An Ensemble method for Content Selection for Data-to-text Generation}

\author{Dimitra Gkatzia and Helen Hastie \\
  Heriot-Watt University \\
  Riccarton, EH14 4AS \\
  Edinburgh \\
  {\tt \{d.gkatzia, h.hastie \}@hw.ac.uk} 
}
  
\date{}

\begin{document}
\maketitle
\begin{abstract}
We present a novel approach for automatic
report generation from time-series data, in
the context of student feedback generation.
Our proposed methodology treats
content selection as a multi-label 
classification (MLC) problem, which takes as input
time-series data (students' learning data) and outputs a summary of these data (feedback). Unlike previous work, this method considers all data simultaneously using ensembles of classifiers, and therefore, it achieves higher accuracy and F-score compared to meaningful baselines. 
\end{abstract}
\section{Introduction}\vspace{-2mm}
Summarisation of time-series data refers to the
task of automatically generating text from variables
whose values change over time. We consider
the task of automatically generating feedback
summaries for students describing their semester-long performance
during the lab of a Computer Science
module. There have been 9 learning factors identified which contribute to students' learning: (1) marks, (2) hours\_studied, (3) understandability, (4) difficulty, (5) deadlines, (6) health\_issues, (7) personal\_issues, (8) lectures\_attended and (9) revision \cite{Gkatzia2013}.

Gkatzia et al.'s analysis \shortcite{Gkatzia2013} showed that there are 4 ways to refer to a learning factor:
\begin{enumerate}[noitemsep,nolistsep]
\item \textbf{\textless trend\textgreater}: describing the trend, 
\item \textbf{\textless weeks\textgreater}: describing what happened at every time stamp, 
\item \textbf{\textless average\textgreater}: mentioning the average, or 
\item \textbf{\textless other\textgreater}: making another general statement. 
\end{enumerate}

The task of content selection for feedback generation can be formulated as a classification task as follows: given a set of 9 learning factors, select the content that is most appropriate to be included in a summary. Content  is represented by templates. A template is defined as a quadruple consisting of an
\textit{id}, a \textit{factor}, a \textit{reference
type} (trend, weeks, average, other) and \textit{surface
text}.

Overall, for all factors there are 29 different templates. There are two decisions that need to be made: (1) 
whether to talk about a factor and (2) in which way to refer to it. 
Instead of dealing with this task in a hierarchical way, where the algorithm will first decide whether to talk 
about a factor and then will decide how to refer to it, our proposed model treats both steps jointly. The proposed method reduces the decision workload by deciding either in which way to talk about a factor, or not to talk about a factor at all.

 \begin{figure}[t!]
\centering
\includegraphics[width=65mm]{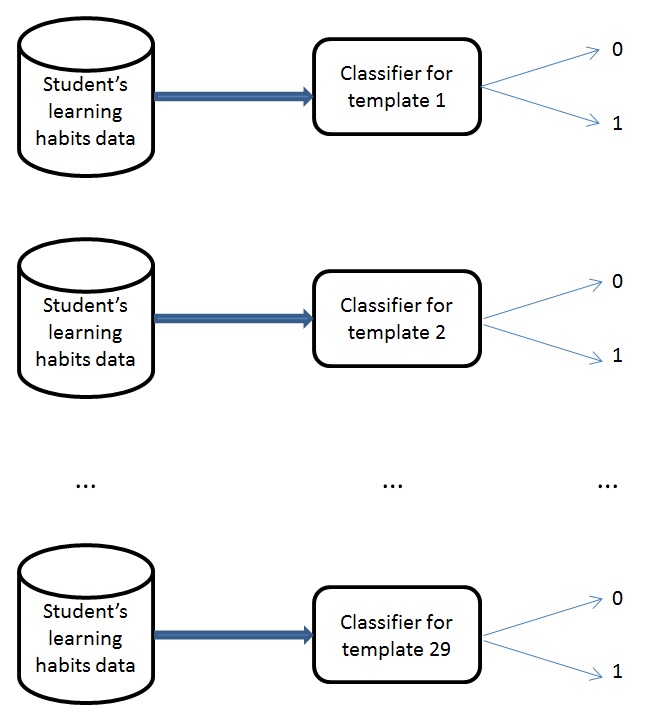}
\caption{Feedback generation as a binary classification problem without history.}
\label{binaryClassif}
\end{figure}
  \begin{figure*}[t!]
\centering
\includegraphics[width=150mm]{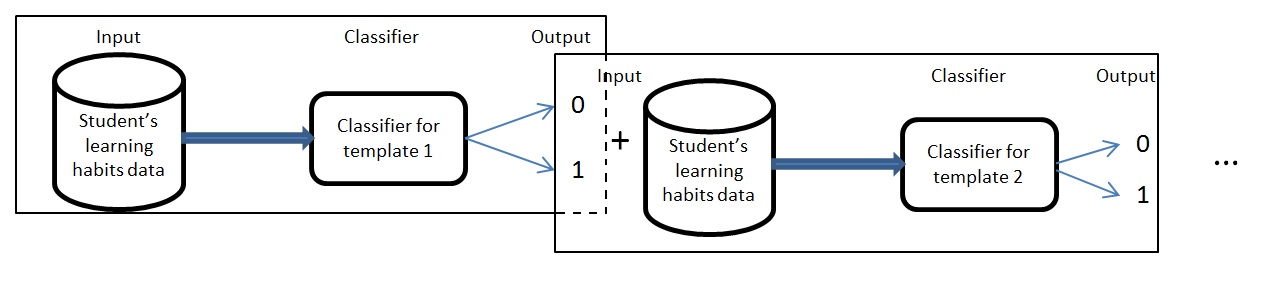}
\caption{Feedback generation as a binary classification problem with history.}
\label{history}
\end{figure*}

\section{Multi-label Classification}\label{multiLabel}\vspace{-2mm}
Classification is concerned with the identification of a category $l$
from a set of disjoint categories $L$ (with $|L| \textgreater 1$) that
an instance belongs to, given the characteristics of the instance. If
$|L| = 2$, then the learning task is called {\it binary
  classification}, for example a task where a classifier is trained to
associate e-mails with either spam or not (i.e. 1 or 0, and hence
binary). If $|L| \textgreater 2$, then the learning task is called
{\it multi-class classification}, for example a task where the
classifier can associate a running area as good, bad or ok. In {\it
  Multi-label} classification (MLC), the instances are associated with
a set of labels $Y \subseteq L$ \cite{Tsoumakas2010}. For example, a
newspaper article can be classified into  {\it health}, {\it science},
{\it economy}, {\it politics}, {\it culture} etc. A specific news
article concerning the breakthrough of the Ebola cure can be
classified into both of the categories {\it health} and {\it
  science}. In the same way, students' data can be assigned labels
that describe them, i.e. each label corresponds to a template. The set of chosen templates can then form a feedback summary.

One set of factor values can result in various
sets of templates as interpreted by the different
experts, i.e. a single student can receive different feedback from
different lecturers. A multi-label classifier is able to make decisions
for all templates simultaneously and capture
these differences. The RAndom k-labELsets
(RAkEL) \cite{Tsoumakas2010} is proposed for tackling content selection.  RAkEL is
based on Label Powerset (LP), a problem transformation
method that uses ensembles of classifiers. LP benefits
from taking into consideration label correlations,
but does not perform well when trained with
few examples \cite{Tsoumakas2010}, as in our case (37 instances). RAkEL overcomes this limitation by constructing
a set of LP classifiers, which are trained
with different random subsets of the set of labels.

\section{Evaluation}\vspace{-1mm}
We compare our approach to four meaningful baselines: 
\textbf{DT (Decision Trees) (no history):} 29 classifiers were trained, each one  responsible for each template. No history is taken into account (see Figure \ref{binaryClassif}).
\textbf{DT (with predicted history):} 29 classifiers were also trained, but this time the input included the previous decisions
made by the previous classifiers (i.e. the history) as well as the set of time-series data in
order to emulate the dependencies in the dataset (see Figure \ref{history}).
\textbf{Majority-class:} It labels each instance with the most
frequent template. 
\textbf {DT (with real history):} A modification of the
previous approach but the real, expert values were used in the model for history rather than the predicted ones.

\section{Results and Conclusions} \label{discussion}\vspace{-2mm}
\begin{table}[t]
\small
\centering
\begin{tabular}{|p{1.9cm}|p{1.29cm}|p{.9cm}|p{.85cm}|p{.6cm}|}
 \hline
\textbf{Classifier }& \textbf{Accuracy (10-fold)}& \textbf{Preci- sion} & \textbf{Recall} & \textbf{F- score} \\
\hline \hline
DT (no history) & *75.95\% & 67.56 & 75.96 &  67.87\\
DT (with predicted history) & **73.43\%  & 65.49& 72.05 & 70.95\\
Majority-class&**72.02\%  & 61.73 & 77.37 & 68.21\\
 \hline
MLC - RAkEL (no history) & \textbf{76.95\%}  & \textbf{85.08} & \textbf{85.94} & \textbf{85.50}\\ \hline
DT (with real history) & \textbf{**78.09\%} & 74.51 & 78.11& 75.54\\
  \hline
\end{tabular}
\caption {Average, precision, recall and F-score of the different classification methods 
(t-test, * denotes significance with $p\textless 0.05$ 
 and ** significance with $p\textless 0.01$, when comparing each
 result to RAkEL.}
\label{comparisonClassif}
\end{table}
MLC - RAkEL \cite{Gkatzia2014} achieves higher accuracy, precision, recall and F-score compared to (1) DT (no history), where each template is predicted from a separate classifier independently, (2) DT (with predicted history), where the decision of the previous template is taken into account in the next decision, similar to Classifier Chains, and (3) a Majority-class baseline.

This method is powerful due to its ability to take into account data correlations \cite{Gkatzia2014}. Multi-label classification should be used when the data to be summarised need to be considered simultaneously and/or when there are limited data available, for example, in student feedback generation, the lectures a student attended is highly correlated with his/her understandability ($r = 0.6$).


\end{document}